\documentclass[runningheads]{llncs}
\usepackage{listings}
\usepackage{xcolor}
\usepackage{float}
\definecolor{codegray}{rgb}{0.5,0.5,0.5}
\definecolor{backcolour}{rgb}{0.97, 0.97, 0.97}
\lstdefinestyle{mystyle}{
    backgroundcolor=\color{backcolour},   
    commentstyle=\color{codegreen},
    keywordstyle=\color{magenta},
    numberstyle=\tiny\color{codegray},
    stringstyle=\color{codegray},
    basicstyle=\ttfamily\footnotesize,
    breakatwhitespace=false,         
    breaklines=true,                 
    captionpos=b,                    
    keepspaces=true,                 
    numbers=left,                    
    numbersep=5pt,                  
    showspaces=false,                
    showstringspaces=false,
    showtabs=false,                  
    tabsize=2
}
\usepackage{graphicx}
\begin{document}
\title{Clinical Trials Ontology Engineering with Large Language Models}
%
%\titlerunning{Abbreviated paper title}
% If the paper title is too long for the running head, you can set
% an abbreviated paper title here
%
\author{Berkan Çakır\orcidID{0009-0006-9922-5091}}
\authorrunning{Çakır et al.}
% First names are abbreviated in the running head.
% If there are more than two authors, 'et al.' is used.
%
\institute{Vrije Universiteit Amsterdam, The Netherlands}
\maketitle              % typeset the header of the contribution
\begin{abstract}
Managing clinical trial information is currently a significant challenge for the medical industry, as traditional methods are both time-consuming and costly. This paper proposes a simple yet effective methodology to extract and integrate clinical trial data in a cost-effective and time-efficient manner. Allowing the medical industry to stay up-to-date with medical developments. Comparing time, cost, and quality of the ontologies created by humans, GPT3.5, GPT4, and Llama3 (8b \& 70b). Findings suggest that large language models (LLM) are a viable option to automate this process both from a cost and time perspective. This study underscores significant implications for medical research where real-time data integration from clinical trials could become the norm.

\keywords{Large Language Model \and chatGPT \and GPT3.5\and GPT4 \and Llama3 \and one-shot \and Ontology \and Ontology Engineering \and Ontology merging \and Clinical Trial \and Diabetes}
\end{abstract}

\section{Introduction}
%introduce not being able to keep up with clinical trials
Clinical trials are essential for the medical industry to give medical practitioners and their patients access to the latest developments in the medical field. However, the number of clinical trials has long outpaced the medical industry's ability to effectively process and incorporate these findings in their practices \cite{bastian_seventy-five_2010}. Recent advancements in transformer-based artificial intelligence (AI) models, such as LLMs, along with the development of ontologies, allow for clinical trials to be processed in an automatic and real-time fashion that can be easily accessed by medical practitioners.

%Broad statement of importance of ontologies
Ontologies have gained significant interest from the academic world \cite{berners-lee_semantic_2001} and industry \cite{chah_ok_2018} due to their ability to link information by reasoning logic, which is not possible with other database structures such as SQL. Ontologies thus allow for "conceptualization" in a shared fashion \cite{guarino_what_2009} by making it relative easy to combine different ontologies. Other benefits of ontologies, particularly when combined with LLM's include alleviating the drawbacks of LLMs through explicit knowledge representation, domain-specific knowledge, and interpret-ability/transparency.

%Broad statement of impact of LLM on ontology engineering
Recent advances in Natural Language Processing (NLP) through the use of LLM models \cite{vaswani_attention_2017} have resulted in models such as BERT \cite{devlin_bert_2018,beltagy2019scibert,rasmy_med-bert_2020}, GPT3 \cite{brown_language_2020}, GPT4 \cite{bubeck2023sparks} and Llama \cite{arxivLLaMAOpen} that allow one to easily process vast amounts of text with just a prompt. These transformer models whilst extremely impressive are inherently random in their responses to prompts. Where the same prompt can result in significantly different responses \cite{bubeck2023sparks}. Transformer models are also prone to give factual incorrect information, also known as "hallucinations" \cite{ji_survey_2023}. Which is mitigated in this study by directly using the outcomes of the clinical trials in the LLM prompt. As well as including an example of how the ontology should look like in the prompt.

%Problem Statement, Research Question, findings + structure
This paper will compare the OQuaRE performance, time, and cost of GPT3.5, GPT4, and Llama3 (8b \& 70b) created ontologies with that of a pre-built golden-standard ontology made by a human layman. Alongside proposing a specialised novel methodology that can be used for ontology merging in the context of clinical trial outcomes with a time complexity of $O(n)$. Findings in Table 1 show significant reduction in both cost and time. Demonstrating practical use of current state-of-the-art LLMs in processing clinical trials in an automated way, with GPT4 in particular showcasing performance near human performance.

%structure
The remainder of this paper is structured as follows: section 2 reviews related work on how simple machine learning techniques are being replaced by LLMs in the context of Natural Language Processing (NLP) and ontology matching. section 3 shows a high-level overview of the methodology, after which describing methodologies and tools used to create and compare the clinical trials. section 4 presents the results, discussing the cost and time differences between the LLMs. Section 5 further discusses these results and places them in broader context. Section 6 concludes the findings by outlining the OQuaRE metrics and performance/cost/time per model. Section 7 concludes the paper with a discussion about the limitations of the study and future work.

\section{Related Work}
%entity relation extraction
Entity and relation extraction, also known as Named Entity Recognition (NER), is a classical NLP task involving identifying and categorizing key elements from text into classes and determining the relationship between these classes. Before the advent of LLMs, simple machine learning (ML) models were one of the few tools in NLP and NER. These ML models were generally based on the bag-of-words principle where each word or n-gram was treated as independent from each other. For example, Kiritchenko et al. \cite{kiritchenko_exact_2010} employed Support Vector Machine (SVM) and Rani et al. \cite{rani_semi-automatic_2017} utilized Latent Semantic Indexing (LSI) and Latent Dirichlet Allocation (LDA) to process clinical trial data in a semi-automatic way. These simpler ML models lacked the complexity needed to fully capture entities themselves and their relations.

%end-to-end
Another disadvantage of the ML models was the isolation of sub-tasks, which allowed for errors to accumulate throughout the data pipeline, degrading performance at each stage. Nye et al. \cite{nye_understanding_2021} have shown that in the context of extracting ICO (intervention, comparator, outcome), an end-to-end NLP model resulted in significantly improved performance. Recent works on extracting information from clinical data have predominantly focused on LLMs. Liu et al. \cite{liu_clinical_2021} have shown that CT-BERT, a fine-tuned version of BERT trained on clinical trials, performs significantly better than previous ML models in clinical trial NLP/NER applications. Giorgi et al. \cite{giorgi_end--end_2019} have looked into using domain-specific LLM, in this case bioBERT, and then fine-tuned the model for their specific use case. Resulting in significant improvement in recall and precision scores for NER tasks. Chen et al. \cite{chen_joint_2020} have shown that for joint-learning of NER and Relation Extraction (RE) in a clinical setting, larger language models perform better than domain-specific language models. However, \cite{lehman_we_2023} has shown that smaller domain-specific models not only outperform large domain-agnostic models, but are also more parameter efficient.
%\textbf{<Work 9>} \cite{jin_pico_2018}.

% Go into details of progress of thesisV3
The progress of simple techniques toward LLMs can also be seen in ontology matching such as OntoChatGPT\cite{Palagin_2023}, MapperGPT\cite{matentzoglu2023mappergpt}, and LLMs4OL\cite{giglou2023llms4ol}.  However, this paper proposes a novel ontology merging methodology that is highly specialised for clinical trial outcomes, arguably simpler than most existing methods.

\section{Methodology}
%intro
This chapter describes the methodology, as is outlined in Figure 1, that used to semi-automate the extraction and integration of clinical trial data into an ontology, using state-of-the-art LLM and a simple novel ontology merging method.

\begin{figure}
\includegraphics[width=\textwidth]{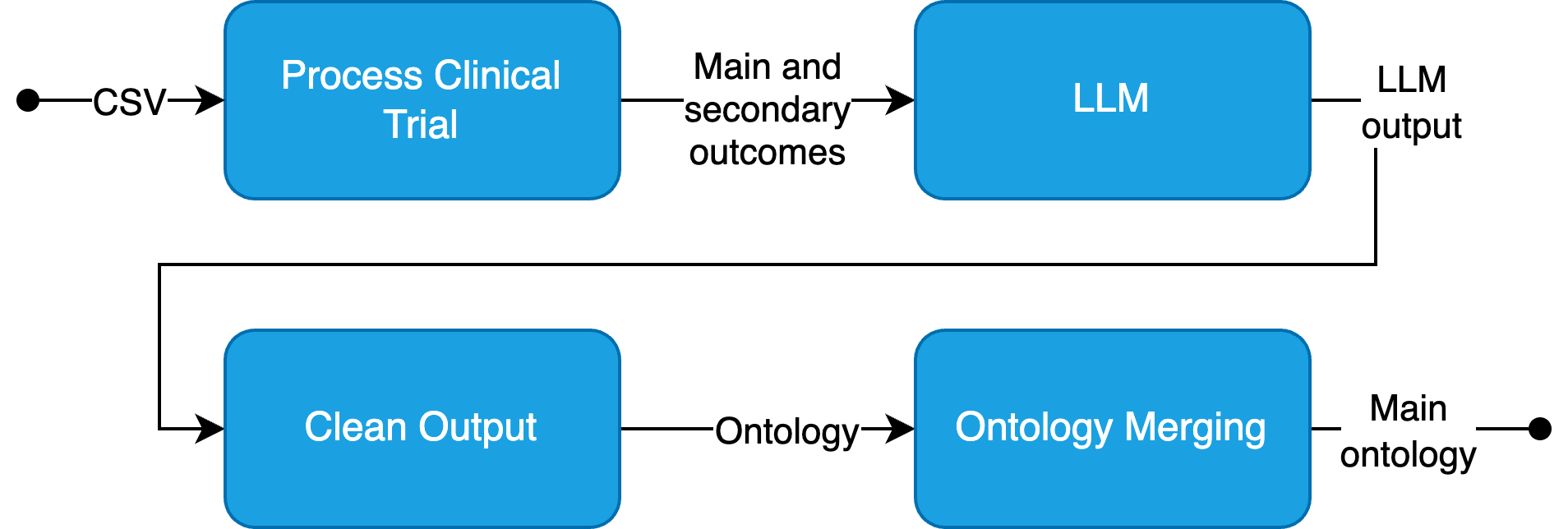}
\caption{High-overview of proposed methodology.} \label{fig1}
\end{figure}

% Explain the pipeline and make a diagram
For the study, 50 clinical trials obtained from the clinicaltrials.gov website were used to obtain the results. The CSV file contains the NCT (unique id for trials submitted on clinicaltrial.gov), primary outcomes, and secondary outcomes. The CSV file containing the clinical trials is fed one clinical trial at a time to be processed. For each clinical trial the LLM is asked to create and ontology based on the primary and secondary outcomes. The individual clinical trial ontologies are then merged through the novel specialised ontology merging methodology to form a single ontology.

\subsection{Dataset}

% Explain the possible issue with rates of the API
% Note that a tiered rate limit applies when using the OpenAI API \cite{openai_rate_limit}. While not relevant for this particular paper. It most likely will become an issue when used at scale.

% clinicaltrial.gov csv and why
The clinical trial data was pulled from the clinicaltrials.gov website in the form of a CSV file. The following filters were applied: "Condition/disease" and "Study Results" which were respectively set to "Diabetes" and "With results". The first 50 clinical trials were then downloaded as a CSV file. It is recommended to keep the disease/condition field set to one condition. As otherwise it will be unclear what outcome measures are related to which condition in the final ontology.

\subsection{LLM models}
% OpenAI API and why
For the LLMs, GPT3.5, GPT4, and Llama3 (8b \& 70b) were selected due to their performance among LLMs \cite{metaMetaLlama}. The GPT models were used through an API supplied by OpenAI. While the Llama3 models were self-hosted through an online hosting service using Ollama \cite{githubGitHubOllamaollama}. With Llama3 8b ran on a single nVidia RTX A4000 (16 GB VRAM) and Llama3 70b on a single nVidia A100 (80GB VRAM).

% Seed
Due to the inherit randomness of transfer models, a seed was used for the GPT models to increase reproducibility. \footnote{Note that as of writing this paper this feature is in BETA and that the OpenAI makes no guarantee that the output will be exactly the same each time.}. Though the API, the "temperature" parameter was set to 0 to keep the output of the LLM as consistent possible. The temperature parameter is not the same as a seed as it uses a "...log probability to automatically increase the temperature until certain thresholds are hit."\cite{openai_temperature}. The Llama3 models however do not provide any parameters to make the output less random.

\subsection{Prompt Engineering}
% prompt engineering
Three different prompt engineering tactics were used to get the most out of the LLM \cite{openai_prompt_engineering}. The tactics used for prompt engineering included:
\begin{enumerate}
  \item Using reference material (i.e. primary and secondary outcomes from the csv file containing the clinical trials).
  \item Writing clear instructions such as asking the LLM to adopt a persona as well as explicitly asking the model to do a (set of) specific tasks.
  \item Providing a reference text which in this case was the base ontology structure that was used for the golden-standard ontology.
  \item Prompt chaining, which breaks the task into multiple steps and prompts by using the output of the first prompt as input for the second prompt.
\end{enumerate}

Both Listing 1.1 and 1.2 use the first three listed tactics, while Listing 1.2 adds prompt chaining to the prompt.

\begin{lstlisting}[language=Python, caption=Prompt engineering code used for ontologies generated without prompt chaining.]
{"role": "system", "content": "You are a computer scientist tasked with creating an ontology from clinical trials in the OWL ontology code format."},
{"role": "assistant", "content": "You have to first extract the biomarkers, endpoint scores, outcome measurement tools, and questionaire types from cinical trial and then turn that into the owl ontology code format" + NCT},
{"role": "assistant", "content": "The only thing you have to do is put the biomarkers, endpoint scores, outcome measurement tools, and questionaires as subclasses of their respective mainclasses (i.e. ex:Biomarker, ex:EndpointScore, ex:MeasurementTool, and ex:Questionnaire)."},
...
{"role": "user", "content": mainOutcomes + " " + secondaryOutcomes}
\end{lstlisting}

\begin{lstlisting}[language=Python, caption=Prompt engineering code used for ontologies generated with prompt chaining.]
{"role": "system", "content": "You are a biologist tasked with extracting biomarkers, endpoint scores, outcome measurement tools, and questionaire types."},
{"role": "user", "content": mainOutcomes + " " + secondaryOutcomes}
...
{"role": "system", "content": "You are a computer scientist tasked with creating a ontology from clinical trials in the OWL ontology code format."},
{"role": "assistant", "content": "You have to convert the biomarkers, endpoint scores, outcome measurement tools, and questionnaire types into an ontology."},
{"role": "assistant", "content": "The only thing you have to do is put the biomarkers, endpoint scores, outcome measurement tools, and questionaires as subclasses of their respective mainclasses (i.e. ex:Biomarker, ex:EndpointScore, ex:MeasurementTool, and ex:Questionnaire)."},
\end{lstlisting}

These prompt engineering tactics each come with their drawbacks. Namely increased cost due to additional token input for the first three listed tactics. As well as increased time and cost with prompt chaining as it requires two separate prompts to be processed.

\subsection{Ontology Merging}
% Cleaning process and why
The output of the LLM has to go through a simple cleaning process. As the model tends to write some additional text before and after the ontology code, even when specifically instructed not to do so. This is however solved programmatically by simply looking at where the ontology code starts and ends and remove part of the output that comes before and after.

% Saving into individual ontologies and why
Due to the token limitation that is inherent of LLMs. The design choice was made to process each clinical trial into separate ontology files to ensure the methodology is theoretically infinitely scalable.

% Ontology merging explain scaling, why (design choices), and how (diagram and clarify each step)
Figure 2 showcases an overview of the ontology merging. The ontology merging stage is needed to add each independent clinical trial ontology to a single main ontology. The ontology merging stage roughly follows the same design as the ontology creation stage. Where the output of the ontology creation stage consisting of independent ontologies is added one by one but at the triple level. One way to do this is by comparing each new triple with all existing triples and asking a LLM if they are conceptionally the same. This approach can be used for small ontologies and is the preferred approach as the LLM can take various factors into account. However, this results into scalability issues when the ontology grows larger with $O(n^2)$. To account for this a sorted synonym list is used. Before a triple is added to the main ontology. The triple and its synonyms are first checked whether they exist in the sorted synonym list. If not, the triple is added and the triple and its synonyms are added to the sorted synonym list. If it does exist, the triple is skipped as it already exists in the ontology. This ensures the lookup necessary for the ontology merging scales with $O(log n)$ instead of $O(n^2)$. The ontology merging stage itself however has a time complexity of $O(n)$ as it loops over triples of all individual clinical trial ontologies one by one.

This particular ontology merging method does come with a major drawback. As the concepts are added one by one, the relations between them are lost. Limiting the applications to an ontology that effectively functions as a categorized list. Resulting in medical practitioners only being able to use the main ontology to check for related concepts in their respective categories for a particular disease. One can opt to use ontology matching instead of the proposed ontology merging method, this however lies outside of the scope of this paper.

% \begin{figure}
% \includegraphics[width=\textwidth]{merge.png}
% \caption{Overview of merging phase.} \label{fig1}
% \end{figure}

\begin{lstlisting}[language=Python, caption=Pseudo-code overview of merging phase.]
Initiate synonymList.
Initiate mainOntology.

For ontology in ontologies:
  If ontology is valid:
    Triple in triples:
      If entity of interest exists in triple:
        Generate synonyms with LLM from entity of interest.
        
        If entity of interest is present in synonymList:
          Add triple to mainOntology
          Add entity and its synonyms to synonymList
\end{lstlisting}

\subsection{Reproducibility}
The used dataset of 50 clinical trials as well as the code used for both generating the individual ontologies and merging of the ontologies used in this study are available via a GitHub repository at https://github.com/berkan-cakir/Clinical-Trials-Ontology-Engineering-with-Large-Language-Models.

\section{Evaluation}
% Two main ways to evaluate, practical (cost, time in ontologyGeneration and ontologyMatching, % of included ontologies, % of classes outside of the four main classes) vs general (OQuaRE)
The evaluation consists of two main parts. First looking at the practical side and comparing the different LLMs and humans with respect to total cost, time to process a clinical trial and transfer the knowledge to an ontology, and percentage of ontologies that were included out of dataset of 50 clinical trials. The second part being a general evaluation using the OQuaRE framework that allows one to compare different ontologies based on 19 different metrics to compare ontology quality in ontologies that are too large for humans to manually compare \cite{DuqueRamos2013}. As well as a brief discussion of the results and its limitations.

\subsection{Practical Evaluation}
Table 1 showcases said practical metrics relevant for extracting outcomes from clinical trials involving ontologies. Comparing the 4 LLMs: GPT3, GPT4, Llama3 (8b), and Llama3 (70b) in addition to using the chaining method for each LLM. The total cost for the GPT models was calculated based the number of tokens provided by the OpenAI API and the model's respective cost per 1000 tokens. The total cost for the LLama models were calculated based on the hourly rate of the rented GPU servers. Time per trial was calculated by averaging total time spent over fifty clinical trial generated by the LLM as well as matching the individual ontology generated from a clinical trial with the main ontology. This percentage reflects whether the generated ontologies had valid syntax and could be included in the main ontology.

\begin{table}[H]
\centering
\caption{Cost and time breakdown per clinical trial (n=50) for each model, human extrapolated from n=14 and assumes hourly wage of \$20. GPT3 model used: gpt-3.5-turbo-1106, GPT4 model used: gpt-4, and LLama3 (8b \& 70b): instruct variant. }\label{tab1}
\begin{tabular}{|l|l|l|l|}
\hline
Model & Total Cost / Trial & Total Time / Trial & Included Ontologies\\
\hline
Human &  $\approx$ \$5 & $\approx$ 15 min. & 100\% \\
GPT3 & \$0.0054 & 143 sec. & 76\% \\
chainedGPT3 & \$0.0072 &  210 sec. & 80\%\\
GPT4 & \$0.0624 &  107 sec. & 26\%\\
chainedGPT4 & \$0.0941 &  212 sec. & 86\%\\
Llama3 (8b) & \$0.0053 &  56 sec. & 28\%\\
chainedLlama3 (8b) & \$0.0082 &  87 sec. & 24\%\\
Llama3 (70b) & \$0.0579 &  110 sec. & 54\%\\
chainedLlama3 (70b) & \$0.0898 &  171 sec. & 74\%\\
\hline
\end{tabular}
\end{table}

\begin{table}[H]
\centering
\caption{Cost en time breakdown of generating individual ontologies from a clinical trial. Ontology generation n=50 for each model, human extrapolated from n=14 and assumes hourly wage of \$20. GPT3 model used: gpt-3.5-turbo-1106, GPT4 model used: gpt-4, and LLama3 (8b \& 70b): instruct variant. }\label{tab1}
\begin{tabular}{|l|l|l|}
\hline
Model & Gen. Cost / Trial & Gen. Time / Trial\\
\hline
Human & $\approx$ \$3.33 & $\approx$ 10 min. \\
GPT3 & \$0.0022 & 43 sec.\\
chainedGPT3 & \$0.0030 &  88 sec.\\
GPT4 & \$0.0594 &  36 sec.\\
chainedGPT4 & \$0.0899 &  47 sec.\\
Llama3 (8b) & \$0.0016 &  17 sec.\\
chainedLlama3 (8b) & \$0.0015 &  16 sec.\\
Llama3 (70b) & \$0.0189 &  36 sec.\\
chainedLlama3 (70b) & \$0.0189 &  36 sec.\\
\hline
\end{tabular}
\end{table}

\begin{table}[H]
\centering
\caption{Cost and time breakdown of ontology merging. Ontology merging n=50 for each model, human extrapolated from n=14 and assumes hourly wage of \$20. For GPT models: gpt-3.5-turbo-1106, and LLama3: instruct variant, 8b and 70b respectively. }\label{tab1}
\begin{tabular}{|l|l|l|l|}
\hline
Model & Merge Cost / Trial & Merge Time / Trial  & Included Ontologies\\
\hline
Human & $\approx$ \$1.67 & $\approx$ 5 min. & 100\% \\
GPT3 & \$0.0033 & 100 sec. & 76\% \\
chainedGPT3 & \$0.0042 &  121 sec. & 80\%\\
GPT4 & \$0.0030 &  71 sec. & 26\%\\
chainedGPT4 & \$0.0043 &  165 sec. & 86\%\\
Llama3 (8b) & \$0.0037 &  39 sec. & 28\%\\
chainedLlama3 (8b) & \$0.0067 &  71 sec. & 24\%\\
Llama3 (70b) & \$0.0389 &  74 sec. & 54\%\\
chainedLlama3 (70b) & \$0.0709 &  135 sec. & 74\%\\
\hline
\end{tabular}
\end{table}

\subsection{OQuaRE Evaluation}
To obtain the OQuaRE results the ontologies were processed in an automated way using an implementation tool called OQuaRE-Metrics \cite{Oquare_tool}. As NOCOnto (Number of Children) is defined as "Mean number of the direct superclasses per class minus the subclasses of Thing"\cite{DuqueRamos2013}. NOCOnto was the only relevant OQuaRE metric due to the ontology merging step omitting relationships as well as the prompt asking to extract a specific set of categories in a specific format. Thus the metric is used as a barometer of concept extraction efficiency.

Figure 2 shows chainedGPT4 being the best performing model. While also showcasing a trend of improved performance of larger models with more parameters. Chain prompting in general seems to have significantly improved entity extraction across all LMMs, with the greatest improvement in models with greater parameters. LLama3 models overall show significant worse performance on the NOCOnto metric than was initially expected. As well as a relatively modest performance boost between chained and non-chained models relative to GPT models. However, taking into account that GPT3.5 has rumoured to have 175b parameters \cite{microsoftComparingGPT35} compared to LLama3's 70b, Llama3 performs surprisingly well compared to GPT3. Figure 3 shows no significant differences between models on any non-NOCOnto metric as expected due to the ontology merging phase.

% goldstandard vs ontologymatched ontologies (GPT3.5 & chained, GPT4 and chained, GPT3.5 fine-tuned & chained, GPT4 fine-tuned and chained)
\begin{figure}
\includegraphics[width=\textwidth]{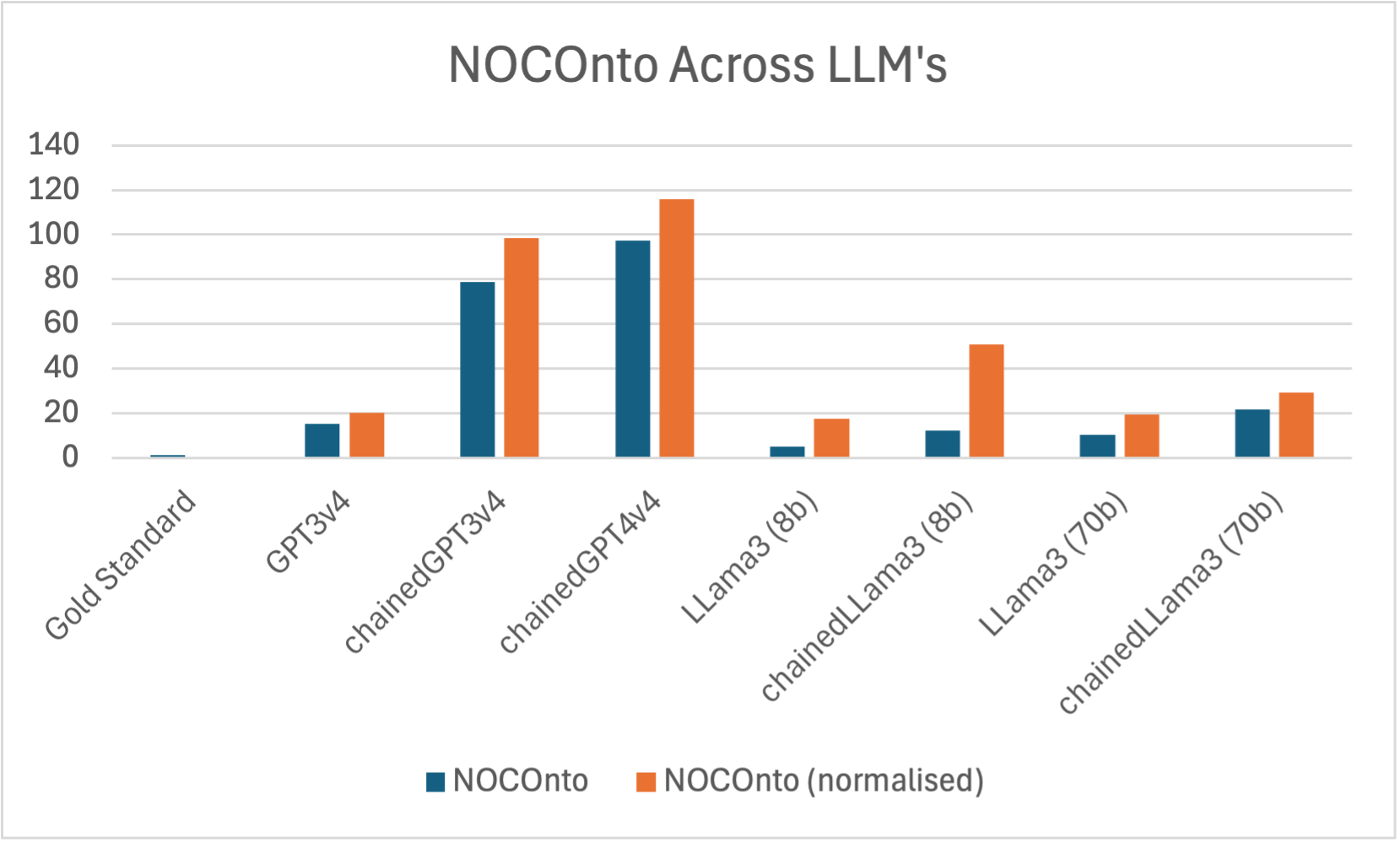}
\caption{NOCOnto metric across different LLMs and or used techniques. NOCOnto(normalised) takes into account the differences in ontologies not being included due to syntax errors (See Table 1, column "included ontologies"). Note that GPT4v4 results are unknown and thus not included.} \label{fig1}
\end{figure}

\begin{figure}
\includegraphics[width=\textwidth]{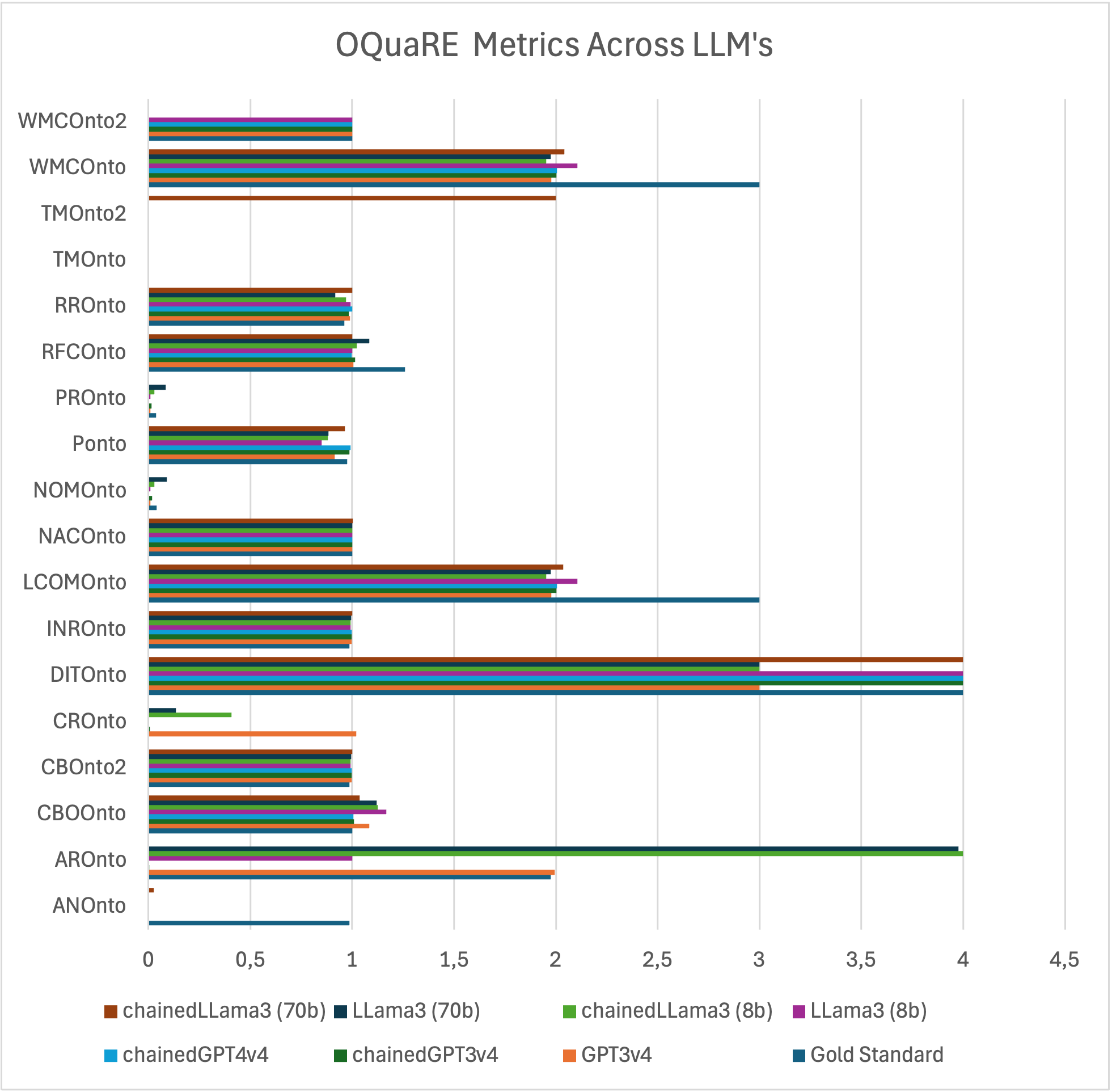}
\caption{All OQuaRE metrics across different LLMs and or used techniques. Note that GPT4v4 results are unknown and thus not included. Also note that NOCOnto OQuaRE metric has been separated due to formatting.} \label{fig1}
\end{figure}

\subsection{Results Discussion}
% Dicuss whether the findings are practically relevant.
\paragraph{Results}
Results indicate that LLMs are both cost-effective and time efficient when compared to manual processing. The only downside being the number of included ontologies at first try. This downside can however be easily fixed when used in production by asking the LLM to generate an ontology of a clinical trial again with a different seed or by using a slightly different prompt. One particular observation that is made is that the ontologies generated by GPT4 is subpar compared to the other models. This seems to be caused due to the high occurance of missing of prefixes in the generated individual ontology, resulting in the ontology not being valid and thus not being included.

\subsection{Limitations of Evaluation}
The evaluation of the performance of the models is exclusively based on the NOCOnto OQuaRE metric, i.e. extracting as many concepts as possible. This metric does not take into account the degree of hallucinations of concepts or the correct categorization of concepts. Another important metric that is not included are the relationships between the concepts, which can be essential in the context of medical applications.

% Discuss the findings by comparing the different OQuaRE scores
%\paragraph{Discussion}
%In Figure 3 one can see that OQuaRE metrics are in general the same compared to all other LLM outputs. A couple of minor exceptions include WMCOnto (Weighted Method Count), LCOMOnto (Lack of Cohesion in Methods). Which results from the focus of the prompt to include biomarkers and the like while omitting any mention of information related to them. A major exception is the NOCOnto metric seen in Figure 3. Which can be explained due to the proportion of clinical trial outcomes were included ontology. As well as the LLMs ability to extract outcomes from clinical trials.

\section{Limitations and Future Work}
\paragraph{Limitations}
This study faced several limitation that may affect the interpretation and generalizability of the results. Firstly, the clinical trials only include diabetes as disease by design. Possibly limiting generalizability to other diseases. Secondly, the sample size of the clinical trials was relatively small (n=50) due to limited financial resources.

% Explain what can be improved such as the rate limit, using open-source models such as LLama2, 
\paragraph{Future Work}
This study has explored the extraction of various medical concepts related to biomarkers, endpoint scores, measurement tools, and questionnaires into sub-classed derived from individual clinical trials. A significant limitation identified is the current ontology merging phase's omission of the relationships between concepts. Which are crucial for practical medical applications. Future research should address this gap by developing methods to accurately preserve and represent the relationships. Further work is needed to mitigate hallucinations

\section{Conclusion}
% Make a conclusion based on scores and relevancy disussion
LLMs seem to be a viable option to process the ever increasing number of clinical trials into an ontology usable by those in the medical field. Extrapolating the practical metrics of chainedGPT4 from 50 clinical trials to 6200 (the number of total clinical trials about diabetes available on clinicaltrials.gov that include outcome measures) results in an estimated total cost of roughly \$584 and taking a little bit more than 15 days to process. Compared to if a human were to process the clinical trials resulting in a cost of roughly \$31000 and taking 1550 hours. Which are roughly 193 eight-hour working days or roughly 39 forty-hour working weeks.

%To increase the usability of the ontology future research could focus on including the NCT of the clinical trial to increase transparency and allow medical staff to double check the LLMs work. As well as including synonyms and linking them and possibly informative annotations that explain concepts that medical staff may not be familiar with yet.

\newpage
% ---- Bibliography ----
%
% BibTeX users should specify bibliography style 'splncs04'.
% References will then be sorted and formatted in the correct style.
%
\bibliographystyle{splncs04}
\bibliography{references}

\end{document}